\def\year{2019}\relax
\documentclass[letterpaper]{article} 
\usepackage{aaai19}  
\usepackage{comment}
\usepackage{times}  
\usepackage{helvet}  
\usepackage{courier}  
\usepackage{url}  
\usepackage{graphicx} 
\usepackage{amsmath}
\usepackage{amsmath,amssymb}
\usepackage{mathrsfs}
\usepackage[normalem]{ulem}
\usepackage{mathptmx}

\begin{document}
%
\title{Optimization of Information-Seeking Dialogue Strategy\\for Argumentation-Based Dialogue System}

\author{
Hisao Katsumi$^1$, 
Takuya Hiraoka$^2$, 
Koichiro Yoshino$^{1,3}$,\\
\bf{\Large{
Kazeto Yamamoto$^2$, 
Shota Motoura$^2$, 
Kunihiko Sadamasa$^2$, 
Satoshi Nakamura$^1$}}\\
$1$ NARA Institute of Science and Technology\\
$2$ NEC Central Research Laboratories\\
$3$ Japan Science and Technology Agency\\
}
%

\newcommand{\myh}{\mathchar`-}

\newcommand{\ky}[1]{\textcolor{orange}{\bf\small #1} }
\newcommand{\ta}[1]{\textcolor{red}{\bf\small [#1 --HROK]} }
\newcommand{\hk}[1]{\textcolor{cyan}{\bf\small [#1 --KTM]} }
\newcommand{\kyc}[1]{\textcolor{blue}{\bf\small [Comment-YSN: #1]} }
\newcommand{\kye}[2]{\textcolor{orange}{\sout{#1} {#2}}} 
\newcommand{\argmax}{\mathop{\rm arg~max}\limits}
\newcommand{\argmin}{\mathop{\rm arg~min}\limits}

\maketitle



\begin{abstract}
Argumentation-based dialogue systems, which can handle and exchange arguments through  dialogue, have been widely researched.
It is required that these systems have sufficient supporting information to argue their claims rationally; 
however, the systems often do not have enough of such information in realistic situations.
One way to fill in the gap is acquiring such missing information from dialogue partners (information-seeking dialogue).
Existing information-seeking dialogue systems are based on handcrafted dialogue strategies that exhaustively examine missing information.
However, the proposed strategies are not specialized in collecting information for constructing rational arguments. 
Moreover, the number of system's inquiry candidates grows in accordance with the size of the argument set that the system deal with.
In this paper, we formalize the process of information-seeking dialogue as Markov decision processes (MDPs) and apply deep reinforcement learning (DRL) for automatically optimizing a dialogue strategy.
By utilizing DRL, our dialogue strategy can successfully minimize objective functions, 
the number of turns it takes for our system to collect necessary information in a dialogue.
We conducted dialogue experiments using two datasets from different domains of argumentative dialogue.
Experimental results show that the proposed formalization based on MDPs works well, and the policy optimized by DRL outperformed existing heuristic dialogue strategies.
\end{abstract}

\section{Introduction}
Argumentation-based dialogue systems are systems that can handle and exchange arguments with their dialogue partner.
These systems have been widely researched because their ability to argue is required in various argumentative situations, such as persuasion and negotiation~\cite{amgoud2000modelling,mcburney2001representing,parsons2002analysis,parsons2003outcomes}.
In general, an argument consists of a claim and its facts that 
support it
~\cite{besnard2014introduction,besnard2014constructing}.
Systems are required to provide rational arguments in such situations with sufficient information that supports their claims.
For example, if a system works as a prosecutor and tries to argue the claim ``The accused is guilty of larceny,'' it is required to provide supporting facts such as ``a witness saw the accused steal the goods.''

The ability to collect facts in order to construct rational arguments is important for argumentation-based dialogue systems because it is not often that the systems have sufficient facts to support their claims beforehand.
One way for the systems to cover for this lack of supporting facts is collecting the required facts from a dialogue partner or third parties through interactions, as human being do.
For examples, if a system tries to argue that ``The accused is guilty of larceny'' but does not have facts that support this claim, 
it can try to collect them by asking the witness questions such as ``Did you see the accused steal something?''
The process of collecting facts can be modeled as \textit{information-seeking dialogue}~\cite{walton1995commitment,fan2015mechanism}.

In some existing research, the use of handcrafted dialogue strategies for information-seeking dialogue systems has been proposed
~\cite{parsons2002analysis,fan2012agent,fan2015mechanism}.
These strategies are designed to examine all facts exhaustively; 
however, facts that support the main claim must be collected quickly because the available time for the systems to collect facts is limited.

Dialogue modeling based on Markov decision processes (MDPs) is widely used and these researches have tried to optimize dialogue strategies by using reinforcement learning (RL) 
~\cite{levin2000stochastic,williams2007partially,misu2012reinforcement,yoshino2015conversational}.
These strategies are defined as mapping functions (policies) from states to actions and are trained to maximize expected future rewards, e.g., successful dialogue tasks and number of steps (turns) in each dialogue.

In this paper, we formalized information-seeking dialogue as MDPs and applied deep reinforcement learning (DRL) to find the optimal policy in dialogues. 
We used a double deep Q-network (DDQN) to find a good policy that can increase the rationality of a system's claim quickly.
We compared our optimized dialogue strategy 
with existing strategies on the basis of two heuristics, 
depth-first search and breadth-first search.
Experimental results show that the proposed method outperformed existing dialogue strategies in several information-seeking dialogue tasks for argumentation-based dialogue systems.

\section{Related Work}
\subsection{Information-Seeking Process in Dialogue}
\label{sec:info-seek}
Information-seeking, which is a kind of argumentation-based dialogue, in particular frequently appears in the real world~\cite{walton1995commitment}.
It is a process of collecting information in which participants in a dialogue exchange questions and answers to collect various facts.
In previous research, heuristic dialogue strategies which exhaustively examine all facts, have been proposed
~\cite{parsons2002analysis,fan2012agent,fan2015mechanism}.
These strategies do not consider the communication overhead cost,
although, in real-world dialogue,
it is important to reduce the number of inquiries because, in general, the available time for information-seeking dialogue is limited.
In addition, it is demanding to construct a strategy for such exhaustive examination based on heuristic rules because the rules usually depends on relations of collected facts or domain-specific knowledge.
%
In contrast, in this paper, we present an automatic optimization method for building a dialogue strategy by considering the communication overhead cost (time-pressure) and the success rate through statistical learning based on RL.

\subsection{Dialogue Strategy Optimization for Argumentation-based Dialogue Systems}
Some previous researchers applied RL to optimize dialogue strategies for argumentation-based dialogue systems: persuasive dialogue systems
~\cite{hadoux2015optimization,rosenfeld2016strategical,alahmari2017reinforcement} and negotiation dialogue systems
~\cite{georgila2011reinforcement,papangelis2015reinforcement}.
They focused on optimizing the dialogue strategy for presenting arguments given in advance; nonetheless, ordinal argumentation-based dialogue systems do not always have sufficient facts to construct rational arguments.
In comparison, our research focuses on optimizing the dialogue strategy for constructing new arguments.
It contributes to developing the ability of the systems to construct such rational arguments adaptively.

\section{Information-Seeking Dialogue for Constructing Rational Arguments}
\label{sec:const-arg}
\begin{figure}[ht]
 \begin{center}
\includegraphics[scale=0.30]{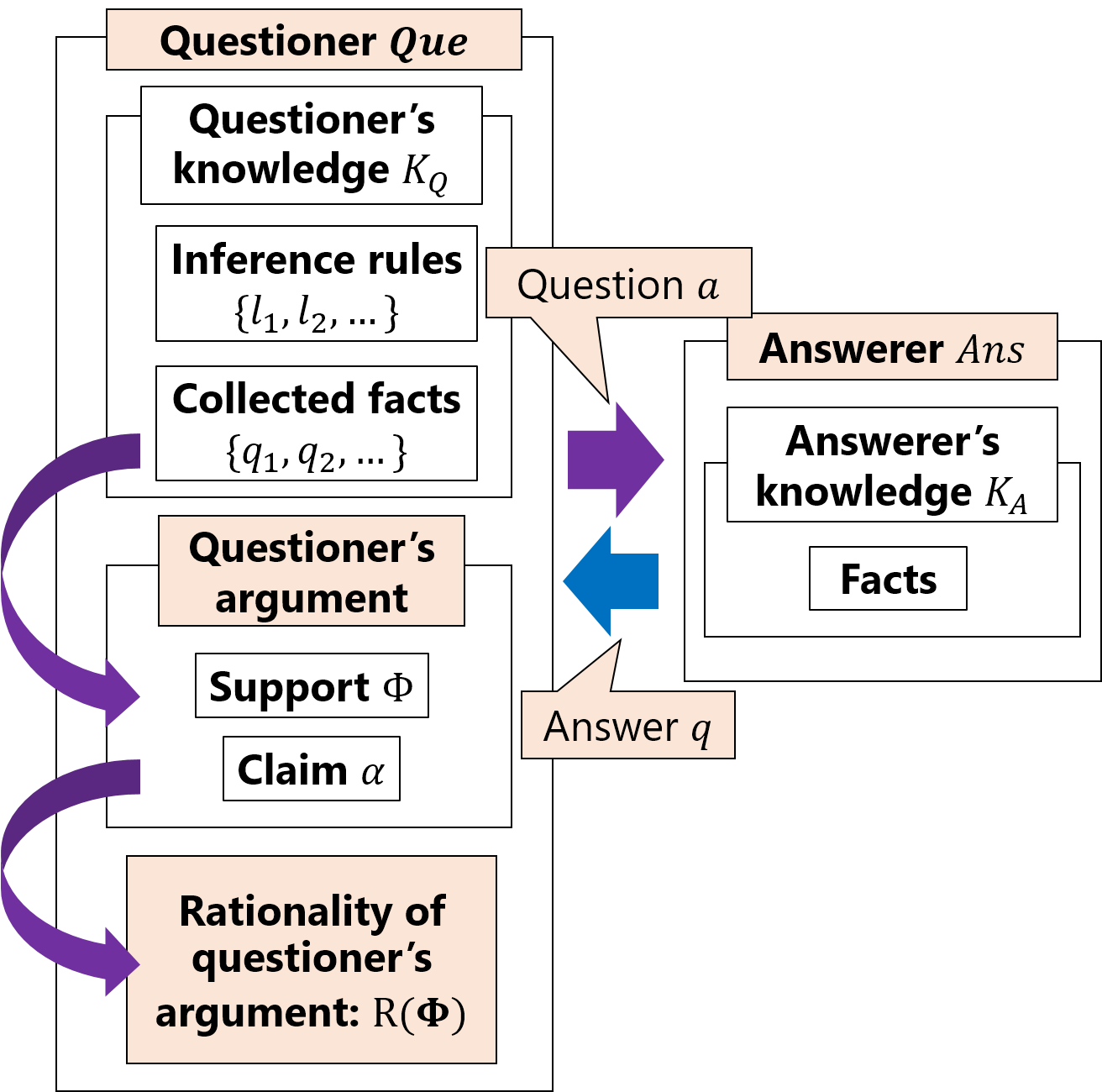}
 \end{center}
\vspace{-5mm}
 \caption{Framework of information-seeking dialogue for constructing rational arguments. }
 \label{fig:dialogeg}
 \vspace{-2mm}
\end{figure}
%
We formulate a formal framework for information-seeking dialogue for constructing rational arguments. 
We introduce a formal definition of arguments, a method for constructing arguments, 
a method for evaluating the rationality of method of the arguments, 
and a dialogue protocol. 

\begin{figure}[ht]
 \begin{center}
\includegraphics[scale=0.35]{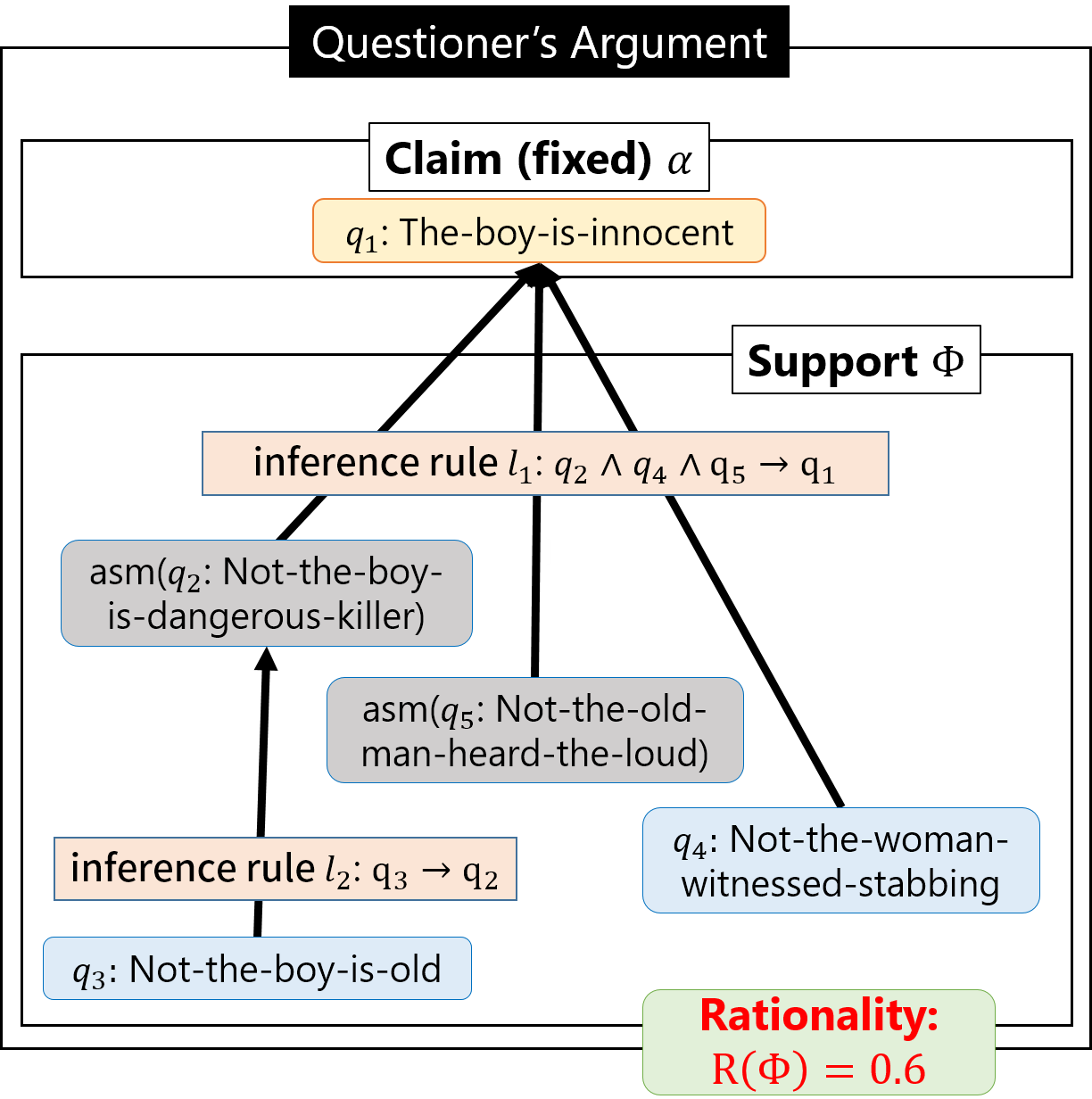}
 \end{center}
 \caption{ Example of questioner's argument. 
 }
 \label{fig:arggeg}
 \vspace{-2mm}
\end{figure}

\subsection{Formal Framework for Information-Seeking Dialogue for Rational Argument Construction}
%
Our proposed dialogue is a kind of an information-seeking dialogue. 
An outline of the information-seeking process is shown in Figure~\ref{fig:dialogeg}.
Questioner $Que$ tries to provide a rational argument including a certain claim but it does not have enough supporting facts. 
In the dialogue, $Que$ asks answerer $Ans$ questions to collect supporting facts. 
$Ans$ provides facts if it knows, i.e., a fact is contained in its own knowledge $K_A$. 
$Que$ updates its knowledge $K_Q$ with the facts that $Ans$ provides. 
The dialogue continues until $Que$ constructs an argument with enough rationality. 

An argument $\langle\Phi,\  \alpha \rangle$ consists of a claim $\alpha$ and its support $\Phi$. 
Here, $\Phi$ is a set of facts and inference rules. 
Formally, 
facts and inference rules are defined by logical expressions. 
Given logical expressions $p_{1}...p_{n},\  q$ satisfying $p_1\land ... \land p_n \rightarrow q$, a fact is represented by a logical expression with $n=0$ (i.e., $q$), and the inference rule is a logical expression with $n\not=0$ (i.e., $p_1 \rightarrow q$). 
Fact $q$ can be substituted with an assumption when the fact is necessary for an argumentation even if it is missing~\cite{dung2009assumption}. 

Figure~\ref{fig:arggeg} shows an example of an argument. 
In this example, $\alpha$ is $q_1$ (yellow rounded box), 
and $\Phi$ is a set of two facts ($q_3$ and $q_4$), two assumptions [$\text{asm}(q_2)$ and $\text{asm}(q_5)$] and two inference rules ($l_1$ and $l_2$).
%
%
%
%
Here, $\text{asm}$ indicates that a fact is an assumption. 
In addition, two inference rules, $l_1$ and $l_2$, are also contained in the argument. 
$l_1$ means that fact $q_1$ is built if we have observed three facts, 
$q_2$, $q_4$, and $q_5$, and 
$l_2$ means that fact $q_3$ is built if we have observed fact $q_2$.
Note that two of the facts, $q_2$ and $q_5$, 
are not contained in $K_Q$, and, thus, assumptions are used instead of them. 

Given a claim $\alpha$ and $Que$'s knowledge $K_Q$,  
the argument $\langle \Phi, \alpha \rangle$ is constructed by defining $\Phi$ as 
$\Phi = K_Q \cup H$, where, $H$ is a set of assumptions. 
$H$ is interpolated as: 
\begin{eqnarray}
&&{\rm arg~min}_{H\in{\mathcal{H}}} \text{E}(\{\alpha\} \cup K_Q \cup H), \nonumber \\ 
&&s.t.,\,K_Q\cup H \models \alpha, \{\alpha\} \cup K_Q\cup H \not\models \bot \nonumber. 
\end{eqnarray}
Here, $\mathcal{H}$ is a set of all possible hypothesis candidates\footnote{$\models$ means that the right term can be derivable from the left term, and $\cup$ means the union of sets. In addition, unnecessary inference rules in $K_Q$ for deriving $\alpha$ are excluded from $\Phi$.}.
$\text{E}$ is a cost function of an abduction model, 
and $\text{E}(\{x, y, \ldots\})$ explains the plausibility of observing $x, y, \ldots$ 
at the same time~\cite{charniak1990probabilistic}.
$E$ is also used in calculating the rationality of the argument $R(\Phi, \alpha)$. 
We will give details on the cost function and rationality in the next section. 

In our proposed dialogue, given $\alpha$, a dialogue proceeds by repeating the following. 
~$Que$ asks $Ans$ about a fact, 
~$Ans$ returns a corresponding fact $q$ if the fact is in $K_A$,
~$Que$ adds $q$ to $K_Q$ ($K_Q \leftarrow K_Q\cup q$), and 
~$Que$ constructs the argument and evaluates its rationality. 
If we process the dialogue with the example of Figure~\ref{fig:arggeg}, 
$\alpha$ is given as $q_1$.
When $Que$ asks $Ans$ 
``Did not the old man, who lived near the scene of the murder, 
hear any loud noises?''
to collect $q_5$, which is necessary to support $\alpha$ for inference rule $l_1$, 
$Ans$ returns $q_5$ as the answer.
Then, $Que$ updates $K_Q$ according to the answer ($K_Q \leftarrow K_Q\cup q_5$). 
Note that only if the corresponding fact to the question, such as $q_5$, is included in $K_A$, $Ans$ return the fact.  
This loop is repeated until the system successfully constructs a rational argument or the number of questions reaches the upper limit.

\subsection{Evaluation Function of Rationality}\label{sec:appx_rationality}
We define the rationality of arguments following an existing piece of work~\cite{ovchinnikova2014abductive}:
\begin{multline}
\text{R}(\Phi,{\alpha}) = (\min_{H'}\text{E}({\{\alpha\} \cup H'}) + \min_{H''}\text{E}(K_Q \cup H''))\\
- \text{E}(\{\alpha\} \cup \Phi).
\end{multline}
We use a cost function that is used in weighted abduction for $\text{E}$ 
(see ~\cite{inoue2011ilp} for more details), and 
it represents the plausibility of arguments of $\text{E}$ occurring at the same time~\cite{charniak1990probabilistic}. 
The smaller $\text{E}(x)$ gets, the more plausible $x$ occurring is; thus, 
equation (1) describes how much $K_Q$ explains the situation, 
where a fact $\alpha$ and those included in $K_Q$ occur at the same time.
To calculate $\text{E}$, 
we use Phillip\footnote{\url{https://github.com/kazeto/phillip}}, an open source fast abductive reasoner. 

\section{Optimization of Information-Seeking Dialogue Strategy based on Reinforcement Learning}
In this section, we first describe the formulation of the information-seeking dialogue for constructing rational arguments, 
on the basis of MDPs; then, we describe how to apply deep reinforcement learning to the formalized information-seeking dialogue to find the optimal strategy efficiently.
\subsection{Formulation based on MDPs}\label{sec:Formalization with MDPs}
In this work, $s_t \in S$, $a_t \in A(s_{t})$ and $r_t \in \mathbb{R}$ denote the internal state of $Que$,
the action taken by $Que$, and 
the reward for $Que$ at time-step $t$, respectively.
We also define 
$\Phi_t$ as a support of the argument at step $t$.

$a_t$ corresponds to the question of $Que$ to $Ans$. 
The action space $A(s_{t})$ at step $t$ is a list of candidates of facts, and $a_t$ is expressed as an index for accessing an element in the list. 
The size of the action space is equal to the number of possible supporting facts, e.g., four in the case of Figure 2. 
In this work, once $Que$ selects $a_t$, $Que$ can not take the same action after step $t$ ($A(s_{t+1}) = (A(s_{t})\backslash \{a_t\})$). 

$s_t$ is composed of an action history record, 
collected facts, and 
the rationality of the argument at step $t$, and 
$s_0$ represents the initial state of the system.
``Action history record'' is the history of actions that $Que$ has taken by step $t$, which is represented as a binary vector $v_{h,t} \in \{0, 1\}^{|A(s_{0})|}$. 
It is necessary for avoiding selecting a question that has been already asked. 
``Collected facts'' 
$\{q_{1}, q_{2},..., q_{t'}\} (t'\leq t)$ are facts that $Que$ has collected from $Ans$. 
We use bag-of-facts (BoF)
to represent collected facts. 
BoF is a representation similar to bag-of-words (BoW) that expresses collected facts with binary vector $v_{q_{t}}$. 
The size of the vector is equal to the number of all possible facts.
As, in the case of Figure~\ref{fig:arggeg}, 
there are four possible facts, 
$q_2$, $q_3$, $q_4$, and $q_5$, the size of the BoF vector is four.
The initial element of each dimension of the vector is set to zero, 
and once a fact is collected, the element corresponding to it is updated to one. 
``Rationality'' at step $t$ is calculated by our rationality function $\text{R}(\Phi_t)$.
The scalar value of the rationality is simply converted to one dimensional vector, $v_{\text{R},t}=[\text{R}(\Phi_t)]$. 
Finally, we concatenate the proposed three features as one vector. 
Thus, the state $s_t$ at time $t$ is defined as: 
\begin{eqnarray}
s_t = [v_{h,t} \oplus v_{q,t} \oplus v_{\text{R},t}].\label{Eqn:feature}
\end{eqnarray}
Note that $\oplus$ represents a concatenation of vectors. 

A reward, $r_t$, is given to encourage $Que$ to collect facts in order to construct more rational arguments as fast as possible. 
Formally, the reward is defined as: 
\[
  r_t = \left\{ \begin{array}{ll}
    r_{\text{time}} + r_{\text{goal}} & (\Theta_{\text{R}} \leq \text{R(}\Phi_{t}\text{)}\\
    r_{\text{time}} & (otherwise)
  \end{array}, \right.
\]
where $r_{\text{time}}$ and $r_{\text{goal}}$ represent a time pressure penalty and 
goal reward, respectively. 
The time pressure penalty is constantly fed as a time-lapse to value dialogues with short steps. 
The rewards are designed to encourage the effective collection of facts for a rational argument. 
A goal reward is fed if $Que$ can construct an argument with higher rationality than a given threshold $\Theta_{\text{R}}$ by the maximum number of dialogue steps ($T_{\text{limit}}$). 

\subsection{Optimization with Double Deep Q-Network}
We consider strategy $\pi(a|s)$ that gives the probability of selecting an action $a \in A(s)$ given a state $s \in S$. 
The objective of reinforcement learning is to find the optimal strategy $\pi^{*}$ that maximizes expected future rewards. 

To find $\pi^{*}$, we use a double deep Q-network (DDQN). 
DDQN~\cite{van2016deep}
is one of the deep RL methods 
based on Q-learning~\cite{watkins1992q}. 
The algorithm introduces double Q-learning~\cite{NIPS20103964} into a deep Q-network~\cite{mnih2015human}. 
DDQN updates the action-value function $Q$ to find $\pi^{*}$.
$Q$ is a function that returns the expected future reward for a pair of $s_t$ and $a_t$. 
DDQN uses a neural network (Q-network) as $Q$. 
Given transitions $\langle s_t, a_t, r_{t+1}, s_{t+1} \rangle$ sampled from the environment, the Q-network is updated to minimize the difference between a target signal, $Y^{\text{DDQN}}_{t}$, and its prediction. 
$Y^{\text{DDQN}}_{t}$ is given as: 
\begin{align}
& Y^{\text{DDQN}}_{t} = r_{t+1} \nonumber\\  
& + \gamma Q(s_{t+1}, \argmax_{a^{\prime}\in A}Q(s_{t+1}, a^{\prime} ;\theta_{t});\theta^{-}_{t}). 
\end{align} 
Here, both $\theta_t$ and $\theta^{-}_t$ are parameters of neural networks at $t$. 
$\theta_t$ is used for both selecting actions and calculating target values, 
while $\theta^{-}_t$ is used only for calculating target values. 
In general, during the learning, 
$\langle s_t, a_t, r_{t+1}, s_{t+1} \rangle$ are collected by agents in accordance with $\varepsilon$-greedy exploration, in which agents choose an action $a_t$ from $A(s_t)$, randomly selecting with probability $\varepsilon$ or by following ${\rm arg~max}_{a_t\in A(s_t)}Q(s_{t}, a_t)$ with probability $1-\varepsilon$.
Once the update of $Q$ converges to the optimal $Q^{*}$, $\pi^{*}$ is given by ${\rm arg~max}_{a_t}Q(s_t,a_t)$. 

When we apply DDQN, it is necessary to build a simulator that outputs the next state $s_{t+1}$ given the current state $s_t$ and the action $a_t$. 
We build an answerer simulator, which returns the required fact if it is included in the knowledge $K_A$; if otherwise, it answers with ``I don't know'' and returns nothing.
$K_A$ is given at the beginning of the dialogue. 

\section{Experiments}\label{Experiments}
We conducted experimental evaluations to investigate the effectiveness of 
dialogue strategies optimized by RL in two different dialogue domains: 
{\it legal discussion} and {\it compliance violation detection}.

We prepared a large number of facts and inference rules 
for $K_A$ and $K_Q$ during training and evaluation
\footnote{
Prepared facts and inference rules for $K_A$ and $K_Q$ are available at: \url{https://github.com/HiKat/Data_for_AAAI-DEEPDIAL19}}. 
In this section, we first describe the prepared 
facts for $K_A$s and inference rules for $K_Q$s and 
then perform experiments with them and discuss the results. 

\subsection{Legal Discussion Domain}\label{sec:Data}
We prepared 550 $K_A$s and 72 inference rules for $K_Q$ generated from an original argumentation dataset, ``Twelve Angry Men dataset~\cite{cabrio2014node}\footnote{\url{http://www-sop.inria.fr/NoDE/NoDE-xml.html#12AngryMen}}."

\begin{figure}[ht]
 \begin{center}
 \includegraphics[scale=0.3]{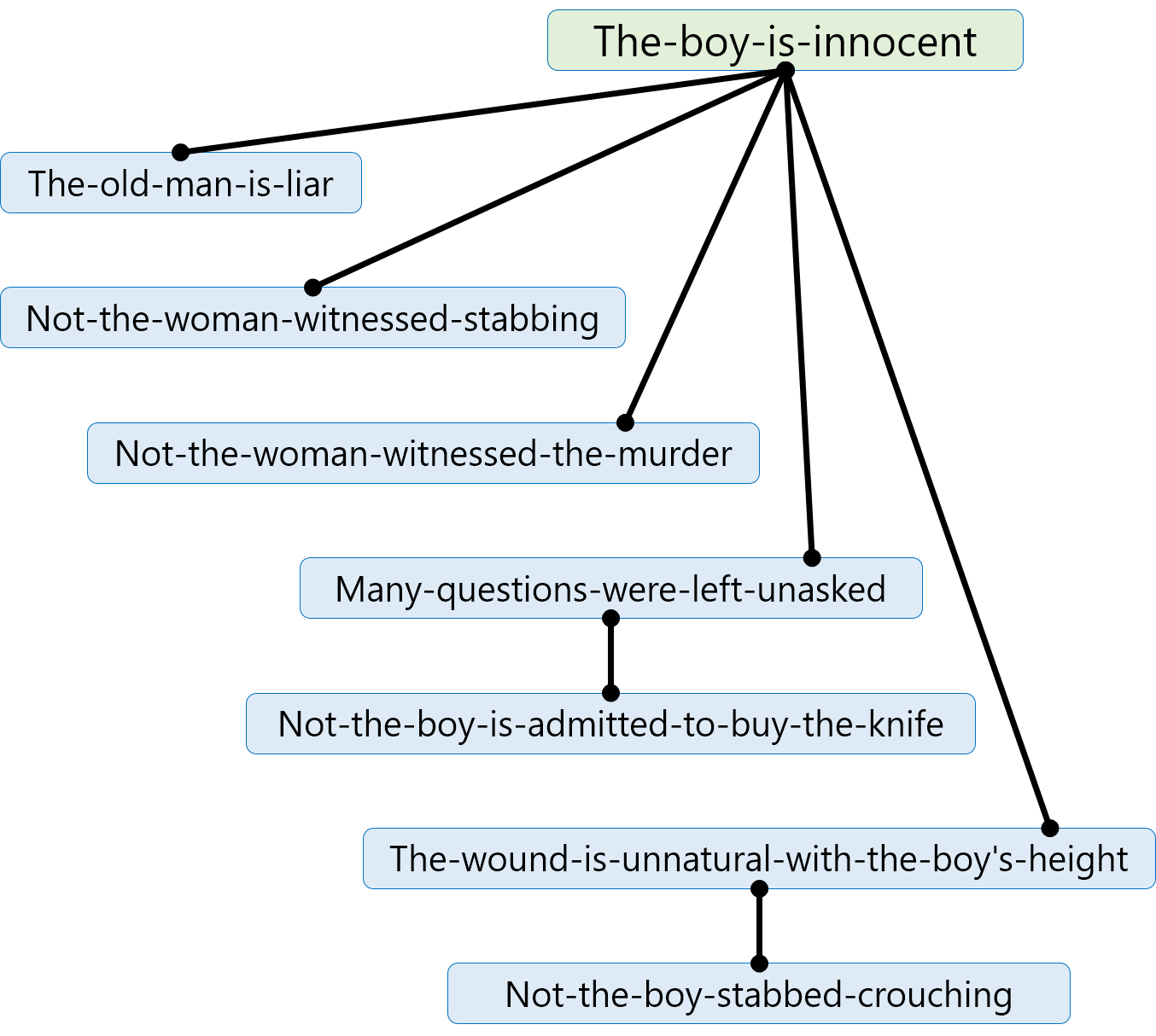}
 \end{center}
\vspace{-5mm}
 \caption{
 Fact graph (FG) made from 122 facts and 72 inference rules 
 extracted from ``Twelve Angry Men dataset"}
 \label{fig:pg_eg}
 \vspace{-2mm}
\end{figure}

This dataset records deliberative arguments (utterances) between the 12 jurors of a homicide trial, and all of the arguments are paired into 80 pairs, 
and each pair is annotated with their relationship: support/attack.
We formally extracted facts and inference rules from the dataset in accordance with the following steps.
\begin{enumerate}
\setlength{\parskip}{0cm} 
\setlength{\itemsep}{0cm} 
\item Consider two utterances as two facts, $\verb|utt1|$ and $\verb|utt2|$ respectively.
\item Consider the relationship between $\verb|utt1|$ and $\verb|utt2|$ as an inference rule, $\verb|utt1| \rightarrow \verb|utt2|$.
\end{enumerate}
As a result, we extracted 160 facts and 80 inference rules by following the procedure above. 
In addition, 
to complicate the information-seeking task,
we combined these extracted rules into more complicated ones.
For example, we combined three rules,
$\verb|utt1| \rightarrow \verb|utt4|$, 
$\verb|utt2| \rightarrow \verb|utt4|$, and
$\verb|utt3| \rightarrow \verb|utt4|$, into one complicated rule,
\begin{eqnarray}
&&\verb|utt1| \wedge \verb|utt2| \wedge \verb|utt3| \rightarrow \verb|utt4|. \nonumber
\end{eqnarray}
Further, removing duplicate facts and inference rules, 
122 facts (including a man claim "The boy is not guilty") and 72 inference rules remained.

All of the 72 inference rules were given to $K_{Q}$ at the beginning of each dialogue, and 121 action candidates corresponding to all facts except the main claim were defined for $Que$.
For the training and evaluation, 
each of 550 $K_A$s was constructed with 20 randomly chosen facts from the 121 facts except the main claim.
500 of the $K_A$s were used for training to find an optimal strategy, 
and the others were used for evaluation.
At the beginning of each dialogue,
the answerer $Ans$ was initialized with one randomly chosen $K_A$, and the $Que$ aimed to construct an argument that claims the fixed fact $\alpha$ ``The boy is not guilty ($\verb|The-boy-is-innocent|$)" by drawing facts repeatedly from $Ans$. 
The rationality $\text{R}(\Phi)$ of $\langle \Phi,\alpha\rangle$ will increase through the information-seeking dialogue if the system has a good dialogue strategy. 

To optimize and evaluate $Que$'s strategy, we prepared a simulator for $Ans$.
The simulator returned one fact corresponding to $Que$'s inquiry if the inquired fact existed in $K_A$.

\begin{table}[ht]
\centering
\caption{
Upper three lines show results for three baseline heuristic strategies, 
random, depth-first search (DFS), and breadth-first search (BFS), strategies, 
and proposed strategy based on DDQN in legal discussion domain.
Lower three lines show results in compliance violation detection domain.
Each cell shows average result of five models generated in parallel and their one standard error.
}
\scalebox{0.85}{
\begin{tabular}{cccc} \hline
                     & \begin{tabular}[c]{@{}c@{}}Average\\ Score\end{tabular}          & \begin{tabular}[c]{@{}c@{}}Completed\\ Episodes\end{tabular}   & \begin{tabular}[c]{@{}c@{}}Average\\ Steps\end{tabular}         \\ \hline \hline
                     \multicolumn{4}{c}{\bf Legal Discussion Domain}\\ \hline
Random               & \begin{tabular}[c]{@{}c@{}}-2.684\\ (1.88)\end{tabular}          & \begin{tabular}[c]{@{}c@{}}3.6\\  (0.92)\end{tabular}           & \begin{tabular}[c]{@{}c@{}}9.884\\ (0.04)\end{tabular}          \\ \hline
Depth-first search   & \begin{tabular}[c]{@{}c@{}}7.176\\ (8.46)\end{tabular}           & \begin{tabular}[c]{@{}c@{}}8.4\\ (4.14)\end{tabular}           & \begin{tabular}[c]{@{}c@{}}9.624\\ (0.18)\end{tabular}          \\ \hline
Breadth-first search & \begin{tabular}[c]{@{}c@{}}0.66\\ (3.30)\end{tabular}            & \begin{tabular}[c]{@{}c@{}}5.2\\ (1.61)\end{tabular}           & \begin{tabular}[c]{@{}c@{}}9.74\\ (0.08)\end{tabular}           \\ \hline
\textbf{DDQN}        & \textbf{\begin{tabular}[c]{@{}c@{}}45.456\\ (5.74)\end{tabular}} & \textbf{\begin{tabular}[c]{@{}c@{}}26.6\\ (2.75)\end{tabular}} & \textbf{\begin{tabular}[c]{@{}c@{}}7.744\\ (0.26)\end{tabular}} \\ \hline \hline
                     \multicolumn{4}{c}{\bf Compliance Violation Detection Domain}\\ \hline
Random               & \begin{tabular}[c]{@{}c@{}}-10\\ (0.00)\end{tabular}             & \begin{tabular}[c]{@{}c@{}}0\\ (0.00)\end{tabular}             & \begin{tabular}[c]{@{}c@{}}-10\\ (0.00)\end{tabular}            \\ \hline
Depth-first search   & \begin{tabular}[c]{@{}c@{}}-10\\ (0.00)\end{tabular}             & \begin{tabular}[c]{@{}c@{}}0\\ (0.00)\end{tabular}             & \begin{tabular}[c]{@{}c@{}}-10\\ (0.00)\end{tabular}            \\ \hline
Breadth-first search & \begin{tabular}[c]{@{}c@{}}-10\\ (0.00)\end{tabular}             & \begin{tabular}[c]{@{}c@{}}0\\ (0.00)\end{tabular}             & \begin{tabular}[c]{@{}c@{}}-10\\ (0.00)\end{tabular}            \\ \hline
\textbf{DDQN}        & \textbf{\begin{tabular}[c]{@{}c@{}}65.304\\ (2.37)\end{tabular}} & \textbf{\begin{tabular}[c]{@{}c@{}}35\\ (1.13)\end{tabular}}   & \textbf{\begin{tabular}[c]{@{}c@{}}4.696\\ (0.12)\end{tabular}}
\\ \hline
\end{tabular}
}
\label{tab:all_test_score}
\end{table}

\begin{table*}[ht]
\caption{
We show example dialogue with strategy optimized with DDQN from evaluation.
Each line of table shows each step of interaction between system and answerer. 
First column shows step number of interaction. 
Second column shows speaker of each step; 
``Sys" stands for system as questioner, and ``Ans" stands for answerer.
Others show each utterance of speaker and rationality of system's argument after collection, respectively.
Answerer's bold-faced utterance corresponds to fact that system successfully collects.
}
%
%
%
\begin{center}
\begin{tabular}{cclc} \hline
Step & Speaker             & Utterance                                                         & Rationality \\ \hline
1    & Sys & Did not the woman across the street see the murder committed?     &             \\
     & Ans           & {\bf The woman across the street did not see the murder committed.}& 0.1         \\ \hline
2    & Sys & Did not the old man hear the kid yell, ``I'm gonna kill you"?     &             \\
     & Ans          & {\bf The old man did not hear the kid yell, ``I'm gonna kill you".}      & 0.4         \\ \hline
3    & Sys & Is the old man a liar?                                           &             \\
     & Ans           & I do not know.                                                    & 0.4         \\ \hline
4    & Sys & Did not the boy crouch down and stab a taller man in the chest?  &             \\
     & Ans            & I do not know.                                                    & 0.4         \\ \hline
5    & Sys & Did not the woman across the street see the boy stab his father? &             \\
     & Ans           & I do not know.                                                    & 0.4         \\ \hline
6    & Sys & Did not the boy admit buying the knife?                          &             \\
     & Ans           & {\bf The boy did not admit buying the knife.}                           & 0.7        \\ \hline
\end{tabular}
\end{center}
%
%
%
\label{Tab:example}
\end{table*}

\subsection{Compliance Violation Detection Domain
} \label{sec:mail-data}
We also prepared 250 $K_A$s, and 106 inference rules as $K_Q$ from a compliance violation detection dataset.
We generated this dataset by gathering thousands of Japanese mail threads from role-played experiments between pairs of subjects.
Each subject played a role as a worker for a certain company or government, 
and he or she made plans together with a partner in compliance violation matters, such as bid-rigging or making a cartel. 
After the role-plays, we extracted 250 mail threads, each of which was composed of about 10 mails.
We generated each $K_A$ from each mail thread by 
converting each mail thread into 100 $\sim$ 600 facts by using Japanese semantic analysis tools 
(ChaPAS\footnote{\url{https://sites.google.com/site/yotarow/chapas}}, 
KNP\footnote{\url{http://nlp.ist.i.kyoto-u.ac.jp/?KNP}} and 
zunda\footnote{\url{https://jmizuno.github.io/zunda/}}) 
removing several facts that has nothing to do with compliance violation.
Finally, each mail thread was converted to 20$\sim$30 facts, and 
3782 kinds of facts remained in all of the 250 threads in total.

The 106 inference rules were manually created to detect compliance violations from each mail thread.
For example, they included the inference rule ``When $x$ is a competitor of $y$ and $x$ provides the price information to $y$, $x$ and $y$ form a cartel.$/\text{competitor(x, y)} \wedge \text{provide\_price\_information(x, y)}\rightarrow\text{cartel(x, y)}$.''
All of the rules were given to $K_{Q}$ at the beginning of each dialogue. 
We used 100 to 250 $K_A$s for training and the others for the evaluation. 

\subsection{Experimental Settings}\label{Methods}
We compared the optimized policy based on RL 
with three baselines strategies based on heuristics in
each of the two domains.
In this section, we give details on the compared strategies and their settings as follows.
{
\setlength{\leftmargini}{0pt} 
\begin{description}
\setlength{\itemsep}{-0.0mm} 
\setlength{\parskip}{-0.0mm} 
\setlength{\itemindent}{0pt}

\item[Random Strategy:]
A random selection strategy. 
The system randomly asks $Ans$ about the existence of facts.
\item[Depth-first Search (DFS) Strategy:]
A heuristic strategy based on depth-first search (DFS). 
$Que$, with this strategy, takes an action in depth-first order across a {\it fact graph} (FG).
In a FG, nodes are facts, and undirected edges are inference rules.
For example, if we have an inference rule, 
$p_1 \wedge p_2 \rightarrow q$, three nodes, $p_1$, $p_2$, and $q$, 
are connected with two undirected edges, $(p_1, q)$ and $(p_2, q)$.
A part of the FG we used in the experiment is shown in Figure~\ref{fig:pg_eg}.
We consider the node corresponding to $Que$'s claim as a {\it claim node}, 
which is the starting node of DFS.
During DFS, $Que$ randomly selects which adjacent node of the current node to ask next.
This strategy is basically the same as the existing work proposed by~\cite{fan2012agent}. 

\item[Breadth-first Search (BFS) Strategy:]
A heuristic strategy based on breadth-first search (BFS).
This strategy also uses the FG as a DFS strategy; 
however, $Que$, with this strategy, takes an action in breadth-first order across the graph.
Note that, $Que$, following this strategy, 
randomly selects one next fact from nodes at the same depth.
\item[DDQN:]
A strategy optimized with DDQN.
In our experiments, the DDQN uses the Q-network, which takes the proposed feature vector (Eqn~\ref{Eqn:feature}) as the input.
The network has two hidden nonlinear layers, 
each of which has 50 units followed by a hyperbolic tangent.
The network finally outputs action-value function values with one linear layer. 
The exploration rate $\varepsilon$ for the $\varepsilon$-greedy strategy was 
linearly annealed from 0.1 to 0.01 during the first 2000 actions.
The discount rate $\gamma$ was fixed to 0.95, and it was trained on 1000 episodes. 
\end{description}
}

In this work, the rewards were set at 
$r_{\text{goal}} = 100$ and
$r_{\text{time}} = -1.0$. 
During training,
$T_{\text{limit}}$ was set to 10.
$\Theta_{\text{R}}$ was fixed to 0.7 for the legal discussion domain and
0.5 for the compliance violation detection domain.

We trained strategies based on DDQN with simulated dialogues, where a simulated $Ans$ played $Que$'s partner. 
At the beginning of each simulated training dialogue, 
the $K_A$ for $Ans$ was randomly selected from 500 or 200 training $K_A$s in response to each domain.
In the evaluation, we also used the simulator; however, 
the $K_A$s for $Ans$ were selected from different 50 $K_A$s in the test set. 
These strategies interacted with the test set simulator for the evaluation.
Note that there was no big difference from conducting a dialogue with humans 
because the operation of $Ans$ is simple in this setting, 
that is only answering regarding the existence of the inquiry.

\subsection{Results}\label{Results}
\begin{figure}[ht]
 \begin{center}
\includegraphics[scale=0.28]{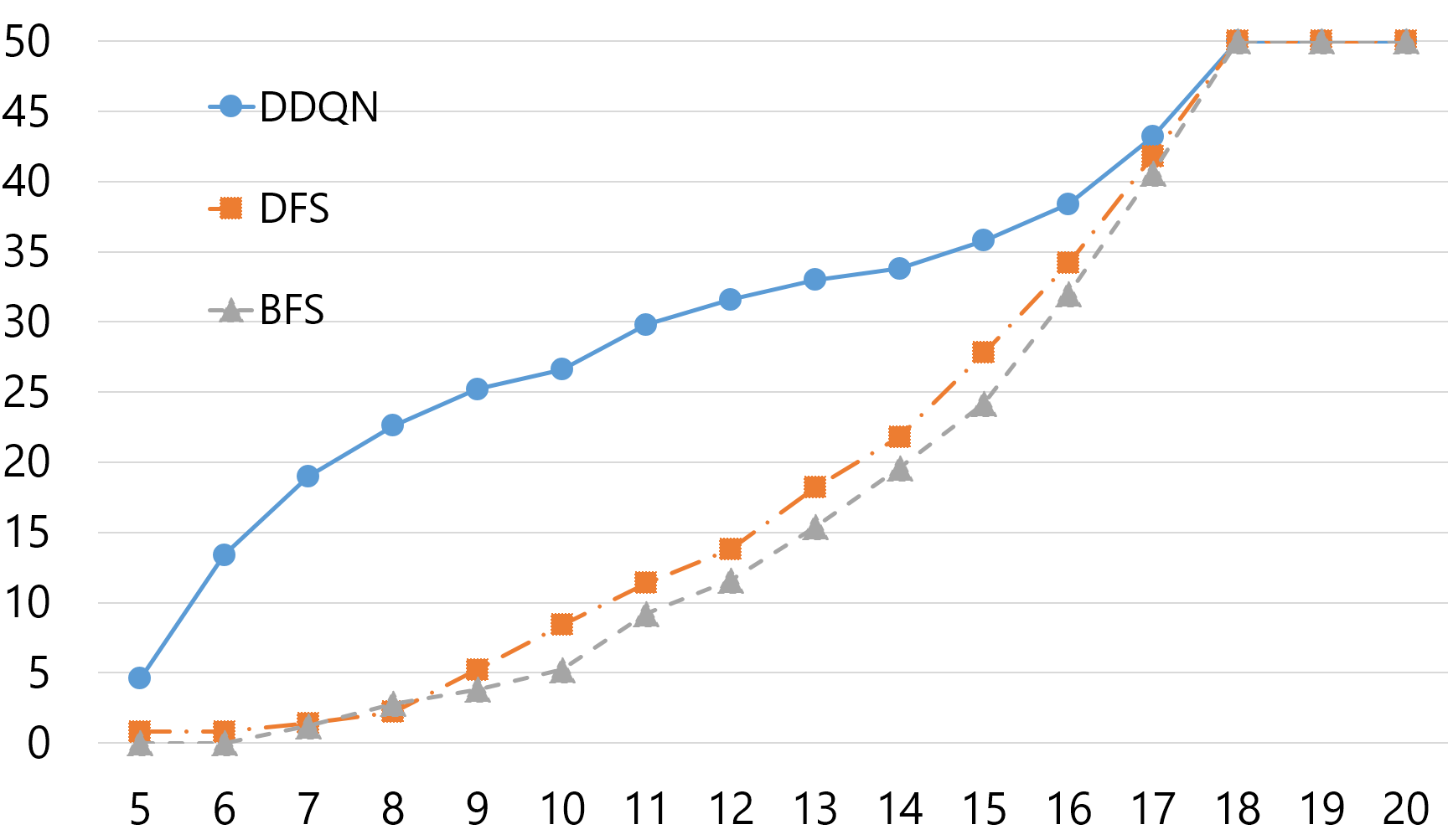}
 \end{center}
\vspace{-5mm}
 \caption{Comparison of number of completed dialogues where questioner constructed rational arguments with each strategy within time limit in legal discussion domain.
 Horizontal axis shows upper limit of questions for each questioner ($T_{limit}$), and 
 vertical axis shows number of completed dialogues. 
 Note that the maximum number of dialogues was 50.
 }
 \label{fig:timelimit-and-completed}
 \vspace{-2mm}
\end{figure}

Table~\ref{tab:all_test_score} shows the scores of the strategies that we compared.
We evaluated strategies in terms of ``test score,'' 
``completed dialogues,'' and ``average steps.''
``Test score'' shows the averages of cumulative rewards at the end of each test dialogue.
``Completed dialogues'' shows the number of evaluated dialogues, in which $Que$ successfully constructed a rational argument whose rationality exceeded the given thresholds within the 10-step limit (i.e., $T_{limit}$). 
``Average steps'' describes the average of the number of steps required to finish dialogues.
The upper part of the table shows the results for the legal discussion domain, and
the lower one shows the results for the compliance violation detection domain.
The results show that the proposed DDQN based strategy achieved the best score in both domains. 
For further analysis, in legal discussion domain, 
we investigated the relationship between the upper time limit of each evaluated dialogue and the number of completed dialogues, as shown in Figure~\ref{fig:timelimit-and-completed}.
This figure indicates that the strategy optimized with DDQN collected facts that support a claim much more quickly than the other handcrafted strategies.

Table~\ref{Tab:example} shows a dialogue example generated by the DDQN strategy.
Each question written in English are prepared by hand with one-to-one mapping from an answer (fact) to a question.
As shown in the example, the questioner uses the strategy based on DDQN to collect the fact ``{\bf The boy did not admit buying the knife}'' at the early stage of the dialogue. 
In the same setting as in Table~\ref{Tab:example}, we observed that both the BFS and DFS strategies collected these facts in a later stage, and, as a result, these strategies required three extra steps to complete the dialogue. 
The aforementioned facts were frequently contained as supports in a rational argument, 
i.e., an argument with higher rationality than a given threshold, 
in these experiments. 
We found that the DDQN strategy tried to collect such frequently contained facts at an early stage 
in many dialogue examples. 

\section{Conclusion}\label{sec:conclusion}
In this paper, we proposed formulating information-seeking dialogue for constructing rational arguments with MDPs.
We also proposed dialogue strategy optimization 
based on DDQN. 
Experimental results showed that strategies optimized with DDQN achieved the best scores in terms of effective information collection to construct rational arguments for the main claims of the system.

As future work, 
considering language understanding errors or the deceptive response of adversarial $Ans$'s are important.
To deal with these problems, we can expand our proposed MDPs formulation to a formulation based on partially observable Markov decision processes (POMDPs). 
Furthermore, we will extend our formulation to 
enriched information-seeking situations, where 
1) $Ans$ can provide arguments based on its own knowledge (inference rules and facts) as well as facts, and 
2) $Que$ can challenge $Ans$ to get further information about the provided arguments. 
\bibliography{AAAI2019}
\bibliographystyle{aaai}

\end{document}